\documentclass[letterpaper]{article} \usepackage[submission]{aaai23}  \usepackage{times}  \usepackage{helvet}  \usepackage{courier}  \usepackage[hyphens]{url}  \usepackage{graphicx}    \usepackage{natbib}  \usepackage{caption}    

\usepackage{epsfig}
\usepackage{graphicx}
\usepackage{hhline}
\usepackage{amssymb}
\usepackage{arydshln}
\usepackage{algpseudocode}
\usepackage{subcaption}
\usepackage{multirow}
\usepackage[flushleft]{threeparttable}
\usepackage{amsfonts}
\usepackage{booktabs}
\usepackage{siunitx}
\usepackage{makecell}
\usepackage{colortbl}

\usepackage{algorithm}
\usepackage{newfloat}
\usepackage{listings}
\DeclareCaptionStyle{ruled}{labelfont=normalfont,labelsep=colon,strut=off} 
\floatstyle{ruled}
\newfloat{listing}{tb}{lst}{}
\floatname{listing}{Listing}
\usepackage{bibentry}

\usepackage{cite}
\usepackage{amsmath}
\usepackage{verbatim}
\usepackage{multirow}
\usepackage{array}
\usepackage{diagbox}[2011/11/22]
\usepackage{mathrsfs}
\usepackage{amsfonts}
\usepackage{mathtools}
\usepackage{multirow}
\usepackage{xcolor}
\usepackage{amssymb}
\usepackage{url}
\usepackage{makecell}
\usepackage{comment}
\usepackage{amsthm}
\usepackage{booktabs}
\usepackage{siunitx}
\usepackage{dsfont}
\usepackage{arydshln}
\usepackage{amssymb,amsmath,amsthm}

\title{Learning Distributions via Monte-Carlo Marginalization}

\author{Chenqiu Zhao}

\newtheorem*{theorem*}{Theorem}

\newtheorem*{lemma*}{Lemma}

\newcommand{\reffig}[1]{Fig. \ref{#1}}

\newcommand{\reftab}[1]{Table \ref{#1}}
\newcommand{\refequ}[1]{Eqn. \ref{#1}}

\newcommand\blfootnote[1]{%
  \begingroup
  \renewcommand\thefootnote{}\footnote{#1}%
  \addtocounter{footnote}{-1}%
  \endgroup
}

\begin{document}

\maketitle

\begin{abstract}
	We propose a novel method to learn intractable distributions from their samples.
	The main idea is to use a parametric distribution model, such as a Gaussian Mixture Model (GMM), to approximate intractable distributions by minimizing the KL-divergence.
	Based on this idea, there are two challenges that need to be addressed.
	First, the computational complexity of KL-divergence is unacceptable when the dimensions of distributions increases.
	The Monte-Carlo Marginalization (MCMarg) is proposed to address this issue.
	The second challenge is the differentiability of the optimization process, since the target distribution is intractable.
	We handle this problem by using Kernel Density Estimation (KDE).
	The proposed approach is a powerful tool to learn complex distributions and the entire process is differentiable.
	Thus, it can be a better substitute of the variational inference in variational auto-encoders (VAE).
	One strong evidence of the benefit of our method is that the distributions learned by the proposed approach can generate better images even based on a pre-trained VAE's decoder.
	Based on this point, we devise a distribution learning auto-encoder which is better than VAE under the same network architecture.
	Experiments on standard dataset and synthetic data demonstrate the efficiency of the proposed approach.
\blfootnote{$*$ authors are equally contributed.}
\end{abstract}

\section{Introduction}
\label{sec_introduction}
Using probabilistic models for generative tasks has become one of the popular areas in machine learning and computer vision \cite{2023_TPAMI_10081412, 2022_TPAMI_9555209}. 
In particular, how to learn the distribution from training data is an interesting problem, since such distributions are usually intractable.
In recent decades, many excellent ideas have been proposed with the Variational Auto-Encoder (VAE) \cite{2014_VAE_Kingma2014} being one of the most popular ones.
The main idea of VAE is introducing a tractable distribution (e.g., a Gaussian) and using variational inference for optimzation. This idea is excellent but possibly imperfect when the output of an encoder does not follow a Gaussian distribution.
For example, when the training data consist of images from two different datasets, such as MNIST \cite{1998_lecun1998mnist} and CelebA \cite{2015_celeba_liu2015faceattributes}, 
the true distribution is not a perfect Gaussian.
Thus, we propose a new question: ``Can we directly learn a parametric distribution model to approximate an intractable distribution, rather than treat the intractable distribution as a tractable one?''
In this paper, the proposed approach can be used to address this question.
\begin{figure}[!t]
    \centering
    \includegraphics[width=0.99\linewidth]{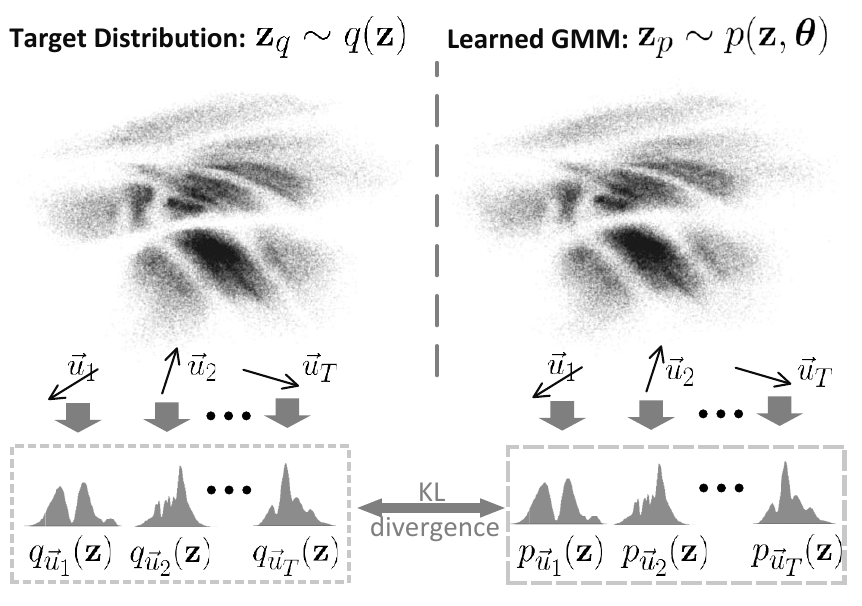}
    \DeclareGraphicsExtensions.
	\caption{Illustration of the proposed approach. We use a Gaussian Mixture Model $p(\mathbf{z};\boldsymbol{\theta})$ to approximate the target distribution $q(\mathbf{z})$,
	by minimizing the KL-divergence between their marginals distributions $q_{\vec{u}}(\mathbf{z})$ and $p_{\vec{u}}(\mathbf{z})$ on different random unit vectors $\vec{u}_1, \vec{u}_2, \cdots \vec{u}_T$.}
    \label{fig_mcidea}		
\end{figure}

The idea of the proposed approach is quite straightforward,
we use a parametric distribution model such as a Gaussian Mixture Model (GMM) to approximate this intractable distribution, 
by using backpropagation algorithms to minimize the KL-divergence between them.
The challenges that need to be addressed can be summarized as:
\begin{itemize}
	\item 1) Dimensionality: When the dimension of a distribution increases, the computational complexity of KL-divergence is too expensive to be acceptable,.
		This challenge is known as the ``curse of dimensionality.''
	\item 2) Differentiability: Since the target distribution is intractable, the probability density function is thus non-differentiable and unavailable. 
		But we intend to use backpropagation algorithms for optimization.
\end{itemize}
In the proposed approach, the Monte-Carlo Marginalization (MCMarg) is proposed to handle the challenge of dimensionality, and Kernel Density Estimation (KDE) is utilized to guarantee the differentiability.

The proposed Monte-Carlo Marginalization is illustrated in \reffig{fig_mcidea}, which is based on a Lemma showing that \emph{identical marginal distributions on all unit vectors imply identical high-dimensional distributions}, please check the method section for a proof.
Based on this Lemma, we do not need to compute the KL-divergence in a high diemnsion space.
Instead, we marginalize the intractable distribution $q(\mathbf{z})$ and the GMM $p(\mathbf{z};\boldsymbol{\theta})$ on a random unit vector $\vec{u}$ to capture their marginal distributions $q_{\vec{u}}(\mathbf{z})$ and $p_{\vec{u}}(\mathbf{z};\theta)$.
By minimizing the KL-divergence of their marginal distributions on different random unit vectors, the GMM will gradually approximate the target intractable distribution.
We call this method Monte-Carlo Marginalization (MCMarg). 
It is also because of the MCMarg method that Kernel Density Estimation (KDE), which is also cursed by dimensionality, can be applied on the samples of the high dimension distribution,
since we only need the marginal distribution.
The samples of intractable distributions are projected on the unit vector $\vec{u}$, then the marginal distribution $q_{\vec{u}}(\mathbf{z})$ can be estimated by KDE.
At the same time, the marginal distribution of the Gaussian mixture model can be easily computed based on its parameters $\boldsymbol{\theta}$ and the projection vector $\vec{u}$, since the Gaussian distribution is a stable distribution.
With enough Gaussian components, almost any intractable distribution can be learned by GMM \cite{goodfellow2016deep}.
Thus, the proposed approach is a powerful tool to handle intractable distributions of the variables in a latent space,
which makes it a better substitute of the variational inference in Variational Auto-Encoders (VAE).
Based on this observation, we propose a Distribution Learning Auto-Encoder (disAE) whose superiority is demonstrated by experiments on strandard datasets, considering identical network architectures.

\section{Related Work} 
Learning distribution information is quite popular and has wide range of applications in computer vision, such as segmentation, classification and generation 
\cite{2021_aaai_Zhou, 2020_aaai_optimal,2020_aaai_policy, 2022_aaai_Liang, zhao2023imbalanced, liu2023boosting, lu2023opt, hung2023reward, qiang2023mixture, yu2023offline}.
However, after a comprehensive review, we found that the interpretations of ``distribution learning'' in different fields are not identical.
For example, the meaning of distribution learning in variational auto-encoder \cite{2014_VAE_Kingma2014} is clearly different than the one in label distribution learning \cite{7890384},
which seems also be different from the proposed approach.
Thus, before discussing related work, we need to clarify the definition of ``distribution learning.''

We follow the definition in \cite{kearns1994learnability} in which ``distribution learning'' is defined as ``learning probability distributions from independent samples.''
It is one of the oldest research topic in computer vision and machine learning.
Based on this definition, we barely find methods that are directly related to the proposed approach, after reviewing literature based on our best effort.
The method closest to ours is the Expectation–Maximization (EM) algorithm \cite{EM_477e7e2b}, since both are designed to update the parameters of GMM based on samples.
However, the EM algorithm involves iteration steps which cannot be incorporated into backpropagation algorithms.
Besides, EM is also suffers from the large number of high dimensional samples in the image generation field \cite{2014_VAE_Kingma2014}.
Finally, the proposed approach is also different from the EM algorithm considering the objective function.
The objective function of the proposed approach is KL-divergence rather than maximizing expectation (EM).
Based on all these reasons, it is difficult to consider that the EM algorithm is directly related to the proposed approach,
even though both methods update the parameters of a GMM.

Normalizing flow \cite{tabak2013family, 2022_PMLR_stimper22a,NEURIPS2022_b6341525} could be another method that is somehow related to the proposed approach, 
since both produce a learned representation of the target distribution.
The main idea of normalizing flow is converting a simple distribution such as the Gaussian into a complex distribution through several steps of invertible transformation, which is usually achieved by a deep learning network \cite{kingma2018glow}.
Unfortunately, when samples from complex distribution are involved, a large number of steps are required. This significantly increases the complexity of the normalizing flow \cite{papamakarios2017masked}.
As a consequence, the computational complexity is very high when it is applied in image generation \cite{dinh2016density}.
In contrast, the proposed approach is intended to learn a GMM to approximate target distrbution. Here only the GMM parameters are involved in backpropagation algorithms, which greatly reduces the computational complexity and memory cost in one learning epoch.

The Vartional Auto-Encoder (VAE) may be another related work, since we also propose a distribution learning Auto-Encoder (disAE).
In order to handle intractable distributions, vartional auto-encoder introduced a recognition model which is assumed to be a multivariate Gaussian distribution \cite{2014_VAE_Kingma2014}.
Based on this assumption, the evidence lower bound is utilized for optimizing the KL-divergence between the encoder output and the Gaussian distribution.
With infinite computational resources, this assumption is nearly perfect and almost becomes the fundamental assumption of modern image generation \cite{2022_TPAMI_9555209} models including the diffusion model \cite{2023_TPAMI_10081412}.
However, considering finite computational resources, 
there is always a gap between the target distribution and the assumed tractable distribution.
One concrete evidence is that the proposed approach is able to find a distribution generating better images based the decoder of a pre-trained VAE.

There are some other methods, such as energy based algorithms \cite{gao2021learning, lecun2006tutorial}, or Markov Chain Monte Carlo \cite{hastings1970monte} methods which are also popular in dealing with distributions or samples. However, either of them can be used to minimize the KL-divergence between intractable distributions and GMM.

\section{Methods}
\label{sec_mcmarg}
\subsubsection{Notation:} Before discussing the details of the proposed approach, the notation used throughout this paper is first introduced.
The main focus of the proposed approach is minimizing the KL-divergence between an intractable distribution and a parametric distribution model, by comparing their marginal distributions on a random unit vector.
We assume that samples of the intractable distribution are available, and Gaussian Mixture Model (GMM) is used as the parametric distribution Model.
Let $\mathbf{z}_q \sim q(\mathbf{z})$ and $\mathbf{z}_p \sim p(\mathbf{z}; \boldsymbol{\theta})$ denote an intractable distribution and the parametric distribution respectively, where $\mathbf{z} \in \mathbb{R}^M$ with size of $M \times 1$,  $\boldsymbol{\theta}$ denote the parameters of $p(\mathbf{z}; \boldsymbol{\theta})$, $q()$ and $p()$ are probability density functions.
Let $\mathbf{Z}=\{\mathbf{z}_q^{(n)} \in \mathbb{R}^{M} |n \in [1, N]\}$ consisting of N independent and identically distributed multivariate samples $\mathbf{z}_q^{(n)}$ from the intractable distribution $q(\mathbf{z})$. Let $q_{\vec{u}}(\mathbf{z}\cdot\vec{u})$ be the marginal distribution of $q(\mathbf{z})$ when we marginalize $q(\mathbf{z})$ on an unit vector $\vec{u} \in \mathbb{U}^M$. Here, $\mathbb{U}^M$ denotes the set of all unit vectors in a $M$ dimensional space. 
Similarly, let $p_{\vec{u}}(\mathbf{z}\cdot\vec{u}; \theta)$ be the marginal distribution when we marginalize $p({\mathbf{z}; \boldsymbol{\theta}})$ on $\vec{u}$,
where $\theta$ denotes the parameters of the marginal distributions that are computed based on $\boldsymbol{\theta}$ and $\vec{u}$ during marginalization.

\subsection{Monte-Carlo Marginalization}
Before discussing differentiability, the challenge of computational complexity with respect to the dimension of a distribution is discussed first.
Mathematically, the KL-divergence between $q(\mathbf{z})$ and $p(\mathbf{z};\boldsymbol{\theta})$ is shown as follows:
\begin{equation}
	D_{KL}(q(\mathbf{z})||p(\mathbf{z};\boldsymbol{\theta}))=\int_{\mathbb{R}^{M}} \!\!\!\!\! q(\mathbf{z}) log \left(\frac{q(\mathbf{z})}{p(\mathbf{z}; \boldsymbol{\theta)}}\right)d\mathbf{z}, 
\end{equation}
where, $\mathbf{z} =\{z_m|m \in [1, M] \cap \mathbb{N}\} = [z_1, z_2, \cdots z_M]^T \in \mathbb{R}^M$ denote vectors of dimension $M \times 1$. 
$q(\mathbf{z})$ is the target distribution which is intractable and  $p(\mathbf{z}, \boldsymbol{\theta})$ is the GMM with $\boldsymbol{\theta}$ as its parameter.
Practically, in order to compute the KL-divergence we need a sequence of discrete values of $z_m$ according to its domain. For example, $z_m = [-1.0, -0.9, \cdots, 0.9, 1.0]$ can be a sequence of the discrete values when $z_m \in [-1, 1]$. In this case, the computational complexity to compute the KL-divergence is $O(21^M)$, which will be impossibly high. 
This phenomenon is known as the ``curse of dimensionality,'' which makes the computation of KL-divergence unacceptable in a high dimension space.

The Monte-Carlo Marginalization (MCMarg) is proposed to handle the computational complexity.
Instead of computing the KL-divergence between $q(\mathbf{z})$ and $p(\mathbf{z};\boldsymbol{\theta})$, we marginalize the target distribution and GMM on random unit vectors to capture their marginal distributions.
Then, by minimizing the KL-divergence of their marginal distributions on all unit vectors, the KL-divergence between the target distribution and GMM is minimized.
Mathematically:
\begin{equation}
	D_{KL}(q(\mathbf{z})||p(\mathbf{z}; \boldsymbol{\theta})) \!\! \Leftrightarrow \!\!\!\! \int_{\vec{u} \in \mathbb{U} } \!\!\!\!\!\!\! D_{KL}(q_{\vec{u}}(\mathbf{z } \cdot \vec{u} )||p_{\vec{u}}(\mathbf{z} \cdot \vec{u};\theta)) d\vec{u}
\end{equation}
where, $q_{\vec{u}}(\mathbf{z})$ and $p_{\vec{u}}(\mathbf{z}; \theta)$ denote the marginal distribution of $q(\mathbf{z})$ and $p(\mathbf{z};\boldsymbol{\theta})$ respectively.
$\vec{u} \in \mathbb{U}^M$ is the random unit vector and $\mathbb{U}$ denotes the set of all unit vectors in a $M$ dimension space.
$\theta$ denotes the parameter of the marginalized GMM, which is computed based on $\boldsymbol{\theta}$ and $\vec{u}$.

Although we cannot perform marginalization on all possible unit vectors for optimization, the Monte-Carlo method can be used to find an approximate solution.
When a sufficient number of random unit vectors are involved in optimization, the GMM will graduately approximate the target distribution.
The proposed Monte-Carlo Marginalization is thus proposed as follows:
\begin{equation}
\begin{aligned}
\mathcal{M}(q(\mathbf{z}), p(\mathbf{z};\boldsymbol{\theta})) & = \int_{\vec{u} \in \mathbb{U} } \!\!\!\!\!\!\! D_{KL}(q_{\vec{u}}(\mathbf{z } \cdot \vec{u} )||p_{\vec{u}}(\mathbf{z} \cdot \vec{u};\theta)) d\vec{u} \\
& \simeq \sum_{\vec{u} \sim \mathbb{U} } \!\! D_{KL}(q_{\vec{u}}(\mathbf{z } \cdot \vec{u} )||p_{\vec{u}}(\mathbf{z} \cdot \vec{u};\theta)),
\end{aligned}
	\label{eqn_mcmarg}
\end{equation}
where, $\mathcal{M}()$ denotes our Monte-Carlo Marginalization method.

The proposed approach can be used to minimize the KL-divergence between $q(\mathbf{z})$ and $p(\mathbf{z}; \boldsymbol{\theta})$ because it is based on an iff (if and only if) condition for the equality between $q(\mathbf{z})$ and $p(\mathbf{z}; \boldsymbol{\theta})$.
In another words, $q(\mathbf{z})$ equals to $p(\mathbf{z}; \boldsymbol{\theta})$ if and only if the marginal distributions of $q_{\vec{u}}(\mathbf{z})$ and $p_{\vec{u}}(\mathbf{z}; \theta)$ are equivalent on any unit vector $\vec{u}$.
Mathematically:
\begin{equation}
\!	\{ \forall \vec{u} \in \mathbb{U}^M \!\!\!\!, q_{\vec{u}}(\mathbf{z} \cdot \vec{u}) = p_{\vec{u}}(\mathbf{z} \cdot \vec{u};\theta) \} \Leftrightarrow \{q(\mathbf{z})= p(\mathbf{z}; \boldsymbol{\theta}) \}
\end{equation}
This conclusion is based on the following Lemma:
\begin{lemma*}
If two multivariate random variables have identical marginal distributions for projections onto all unit vectors, they have identical distributions.
\begin{proof} 
	This lemma is an extension of the ``Uniqueness Theorem for Characteristic Functions'' which has been proved in probability theory \cite{feller1991introduction}.
	Based on it, we just need to prove that the characteristic functions of two distributions are identical.
	This is because the characteristic function contains all the distribution information of the random variable. If the characteristic functions of two random variables are equal, then their distributions must be equal.
	For the convenience of readers, we directly use $q(\mathbf{z})$ and $p(\mathbf{z};\boldsymbol{\theta})$ as examples for the proof.
	Assume two high dimensional random variables $\mathbf{z}_q \sim q(\mathbf{z})$ and $\mathbf{z}_p \sim p(\mathbf{z}; \boldsymbol{\theta})$,
	the characteristic functions of their distributions are shown as:
	\begin{equation}
			\psi_{\mathbf{z}_q }(t)  = \mathbb{E}[e^{-i(t \cdot \mathbf{z}_q )}], \ \ \ \  \phi_{\mathbf{z}_p }(t)  = \mathbb{E}[e^{-i(t \cdot \mathbf{z}_p )}],
	\end{equation}
	where, $t, \mathbf{z}_q, \mathbf{z}_p \in \mathbb{R}^M$.  $i$ is an imaginary number, $e$ is a natural number and $\mathbb{E}$ is the expectation operator. $\psi_{\mathbf{z}_q }(t)$ and $\phi_{\mathbf{z}_p }(t)$ are the characteristic functions of $\mathbf{z}_q$ and $\mathbf{z}_p$ respectively.
	Next, assume we have a unit vector $\vec{u}$ used for projection to perform marginalization.
	The characteristic functions of the marginalized variables $\mathbf{z}_q \cdot \vec{u}$ and $\mathbf{z}_p \cdot \vec{u}$ can be computed as:
	\begin{equation}
			\psi_{\mathbf{z}_q \cdot \vec{u} }(t) = \mathbb{E}[e^{it(\mathbf{z}_q \cdot \vec{u})}], 			\phi_{\mathbf{z}_p \cdot \vec{u} }(t) = \mathbb{E}[e^{it(\mathbf{z}_p \cdot \vec{u})}],
	\end{equation}
	where, $t \in \mathbb{R}$. We can observe that if we have $\psi_{\mathbf{z}_q \cdot \vec{u} }(t) =\phi_{\mathbf{z}_p \cdot \vec{u} }(t) $  for all possible unit vectors $\vec{u}$, this implies that $\mathbf{z}_q$ and $\mathbf{z}_p$ have the same distribution, which means $q(\mathbf{z})$ and $p(\mathbf{z}; \boldsymbol{\theta})$ are identical. The lemma is thus proved.
\end{proof}
\end{lemma*}
\begin{figure*}[!t]
    \centering
	\includegraphics[width=0.99\linewidth]{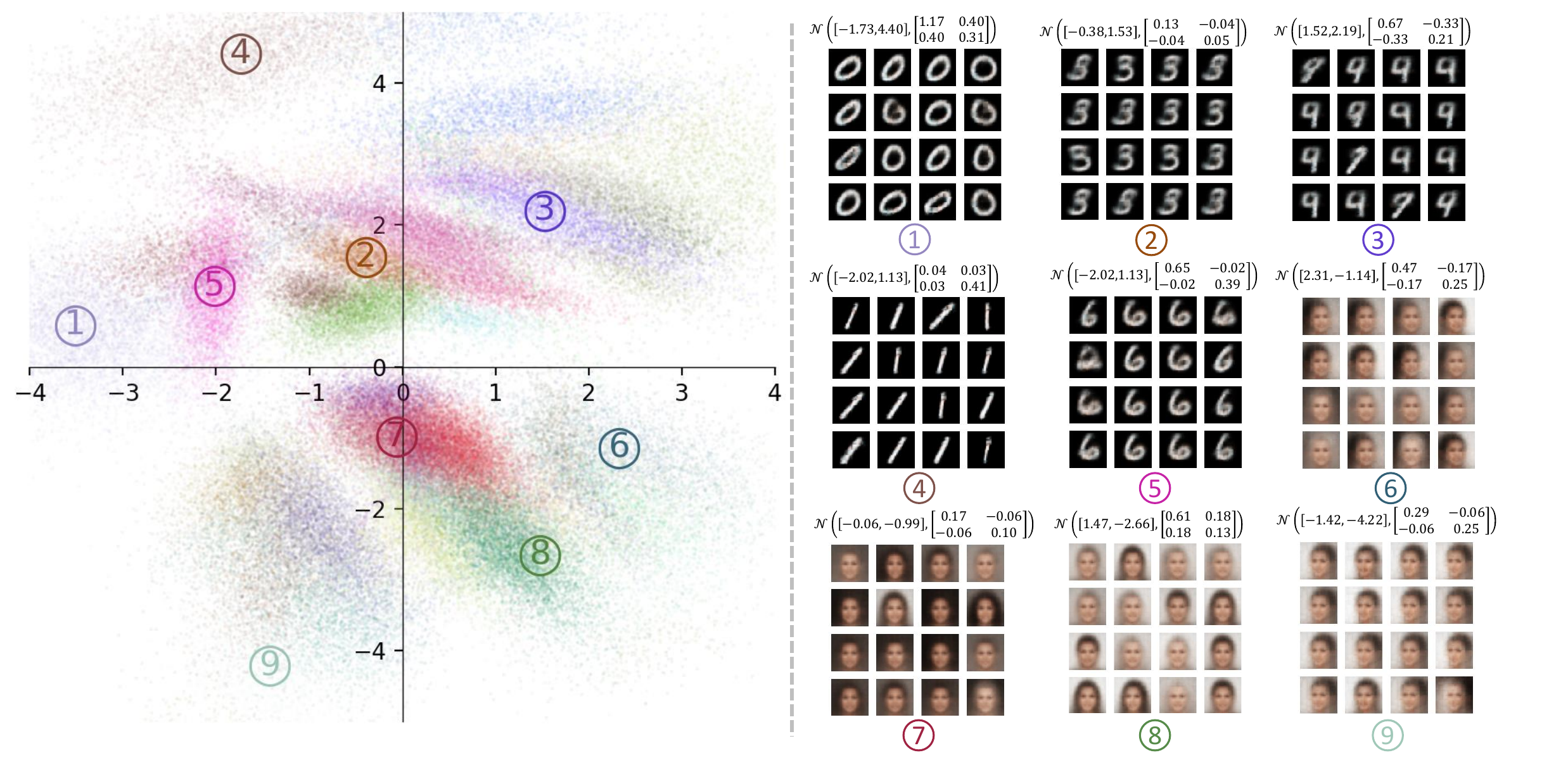}
    \DeclareGraphicsExtensions.
	\caption{The proposed distribution learning Auto-Encoder (disAE) is able to perform self-classification, since a Gaussian mixture model is used to learn and restrict the distribution of latent space variables. We train a disAE with 2d latent space and print the samples from different Gaussian components with different colors. The reconstructed images based on samples from 9 Gaussian components with the highest weights are also illustrated.}
    \label{fig_selfclass}		
\end{figure*}

\subsection{Kernel Density Esitmation and Gaussian Mixture Marginalization}
The next challenge is about differentiability of the optimzation process.
Since the target distribution is intractable, it is reasonble to utilize Kernel Density Estimation (KDE) to capture the probability density function of the target distribution.
Moreover, benefitting from the proposed Monte-Carlo Marginalization, we just need to estimate the marginal distribution based on the projected samples, which greatly reduces the computational complexity in one optimzation epoch.
Assuming that we have the samples $\mathbf{z}_q^{(n)}$ from the target intractable distribution $q(\mathbf{z})$, the approximate estimation of the marginal distribution $q_{\vec{u}}(\mathbf{z}\cdot \vec{u})$ is shown as follows:
\begin{equation}
	q_{\vec{u}}(\mathbf{z}\cdot \vec{u}) \approx  \hat{q}_{\vec{u}}(\mathbf{z}\cdot \vec{u})=\frac{1}{N h}\sum\limits_{ n = 1}^{N} \mathcal{K}\left(\frac{\mathbf{z}\cdot \vec{u}-\mathbf{z}_q^{(n)}}{h}\right)
	\label{eqn_kde}
\end{equation}
where, $\hat{q}_{\vec{u}}(\mathbf{z} \cdot \vec{u})$ denote the esitmation of $q_{\vec{u}}(\mathbf{z})$. $\mathbf{z}_q^{(n)}$ represent the samples of the target distribution $q(\mathbf{z})$ and $N$ is the total number of samples.
$h$ is the bandwidth of the kernel density function, we use $h=0.1$ in our experiment. $\mathcal{K}$ is the kernel function; with commonly the normal distribution function being used.
\begin{figure*}[!t]
    \centering
	\includegraphics[width=0.99\linewidth]{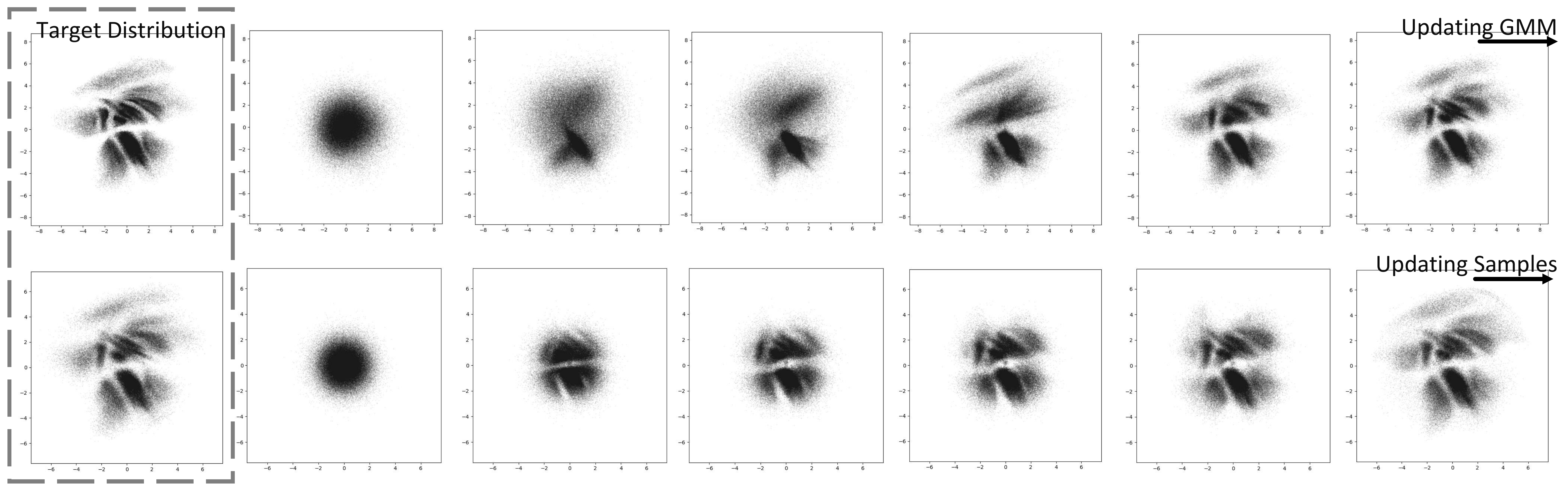}
    \DeclareGraphicsExtensions.
	\caption{The ablation study on the differentiability of the proposed approach. In the first row, the parameters of GMM are updated based on the samples of the target distribution. In the second row, the values of samples are updated based on the target distribution approximated by the learned GMM. }
    \label{fig_bidirect}		
\end{figure*}

As we mentioned above, with enough components, a Gaussian Mixture Model (GMM) can approximate almost any distribution. This is one of the main reasons we use GMM as our parametric distribution model. Besides, another reason is that the Gaussian distribution is stable, with a marginal distribution that is still a Gaussian distribution.
Thus, the marginal distribution of $p_{\vec{u}}(\mathbf{z};\boldsymbol{\theta})$  can be easily computed based on the projection vector $\vec{u}$ and parameter $\boldsymbol{\theta}$.
Assume that the expression of GMM $p(\mathbf{z};\boldsymbol{\theta})$ is shown as follows:
\begin{equation}
	p(\mathbf{z}, \boldsymbol{\theta}) = \sum_{i=1}^{K} w_k \cdot \mathcal{N}(\mathbf{z}; \boldsymbol{\mu}_k, \Sigma_k), 
	\label{eqn_gmm}
\end{equation}
where, $\boldsymbol{\theta}=\{w_k, \boldsymbol{\mu}_k, \Sigma_k \}$ denote its parameter. $\mathcal{N}$() is the Gaussian function. $\boldsymbol{\mu}_k \in \mathbb{R}^M$ and $\Sigma_k \in \mathbb{R}^{M \times M}$ denote the mean and covariance of the $k$-th Gaussian component. $w_k$ denotes the weights. $K$ is the number of components.

Based on the expression in \refequ{eqn_gmm}, the marginal distribution $p_{\vec{u}}(\mathbf{z}\cdot \vec{u}; \theta)$ can be captured as follows:
\begin{equation}
	\begin{aligned}
		p_{\vec{u}}(\mathbf{z}\cdot \vec{u}; \theta) & = \sum_{k=1}^{K} w_k \cdot \mathcal{N}(\mathbf{z}\cdot \vec{u}; \mu_k, \sigma_k^2 ) \\
		& = \sum_{k=1}^{K} w_k \cdot \mathcal{N}(\mathbf{z}\cdot \vec{u}; \boldsymbol{\mu}_k \cdot \vec{u}, \vec{u}^T \Sigma_k\vec{u} ),\\
		\because \mu_k & = \mathbb{E}(\mathbf{z}\cdot\vec{u})  =  \boldsymbol{\mu}_k \cdot\vec{u}  \\
  \sigma_k^2 & = \text{Var}(\mathbf{z}\cdot \vec{u}) =\vec{u}^T \Sigma_k\vec{u}
	\end{aligned}
	\label{eqn_gmm_marg}
\end{equation}
where, $\mu_k$ and $\sigma_k$ denote the mean and standard deviation of the marginal distribution.

Utilizing \refequ{eqn_mcmarg}, \refequ{eqn_kde} and \refequ{eqn_gmm_marg}, we are able to optimize the paramter $\boldsymbol{\theta}$ of GMM $p(\mathbf{z}; \boldsymbol{\theta})$ to approximate target distribution $q(\mathbf{z})$ based on its samples $\mathbf{z}_q^{(n)} \in \mathbf{Z}$.
In addition, since the entire process is differentiable, the proposed approach can be implemented in Pytorch \cite{paszke2019pytorch} with backpropagation algorithms such as Adam \cite{kingma2014adam} being used for optimization.

\subsection{Distribution Learning Auto-Encoder} 
The proposed approach is a very useful tool for learning intractable distributions, which can be a substitute to variational inference in variational auto-encoders.
In order to demonstrate this, we propose the Distribution Learning Auto-Encoder (DisAE) to compare with Vartional Auto-Encoder (VAE),
and demonstrate the efficiency of the proposed approach.

Similar to VAE, our disAE is also a typical auto-encoder architecture. However, with our Monte-Carlo Marginalization (MCMarg) learns the distribution of the latent space.
The images are input into ther encoder to capture the samples of the latent space, which are used as the input of the decoder for reconstruction. 
At the same time, the samples in the latent space are also input into the proposed approach to learn the parameters of the Gaussian Mixture Model. Mathematically:
\begin{equation}
\mathcal{L}(\boldsymbol{\theta}, \alpha , \beta ;\mathbf{x})=\mathcal{M}(q(\mathbf{z}|\mathbf{x} ),p(\mathbf{z};\boldsymbol{\theta}))  + \mathbb{E}_{q_{\alpha }(\mathbf{z}|\mathbf{x}) }\text{log}(g_{\beta }(\mathbf{x}|\mathbf{z})
\end{equation}
where, $\mathcal{L}(\boldsymbol{\theta}, \alpha, \beta; \mathbf{x})$ is our maximum likelihood function. $\mathbf{x}$ denotes the input images and $\mathbf{z}$ denotes its feature vectors in the latent space.
$q_{\alpha}(\mathbf{z}|\mathbf{x})$ and  $g_{\beta}(\mathbf{x}|\mathbf{z})$ denote the encoder and decoder respectively, where $\alpha$ and $\beta$ are their parameters.
$p(\mathbf{z};\boldsymbol{\theta})$ denotes the Gaussian mixture model with $\boldsymbol{\theta}$ as its parameter.

During the training of the proposed approach, the gradient of the latent space variable $\mathbf{z}$ is restricted by GMM, which is also a clustering method.
Thus, the proposed disAE can perform unsupervised classification, as shown in \reffig{fig_selfclass}.
In \reffig{fig_selfclass}, we train our disAE on images from MNIST and CelebA with 2D latent dimension for illustration.
In particular, 128 Gaussian components are used.
We illustrate samples from the 9 Gaussian components with the highest weights, along with the reconstructed images of these samples.
It is clear that the reconstructed images have been self-classified.
\begin{figure*}[!t]
    \centering
    \includegraphics[width=0.99\linewidth]{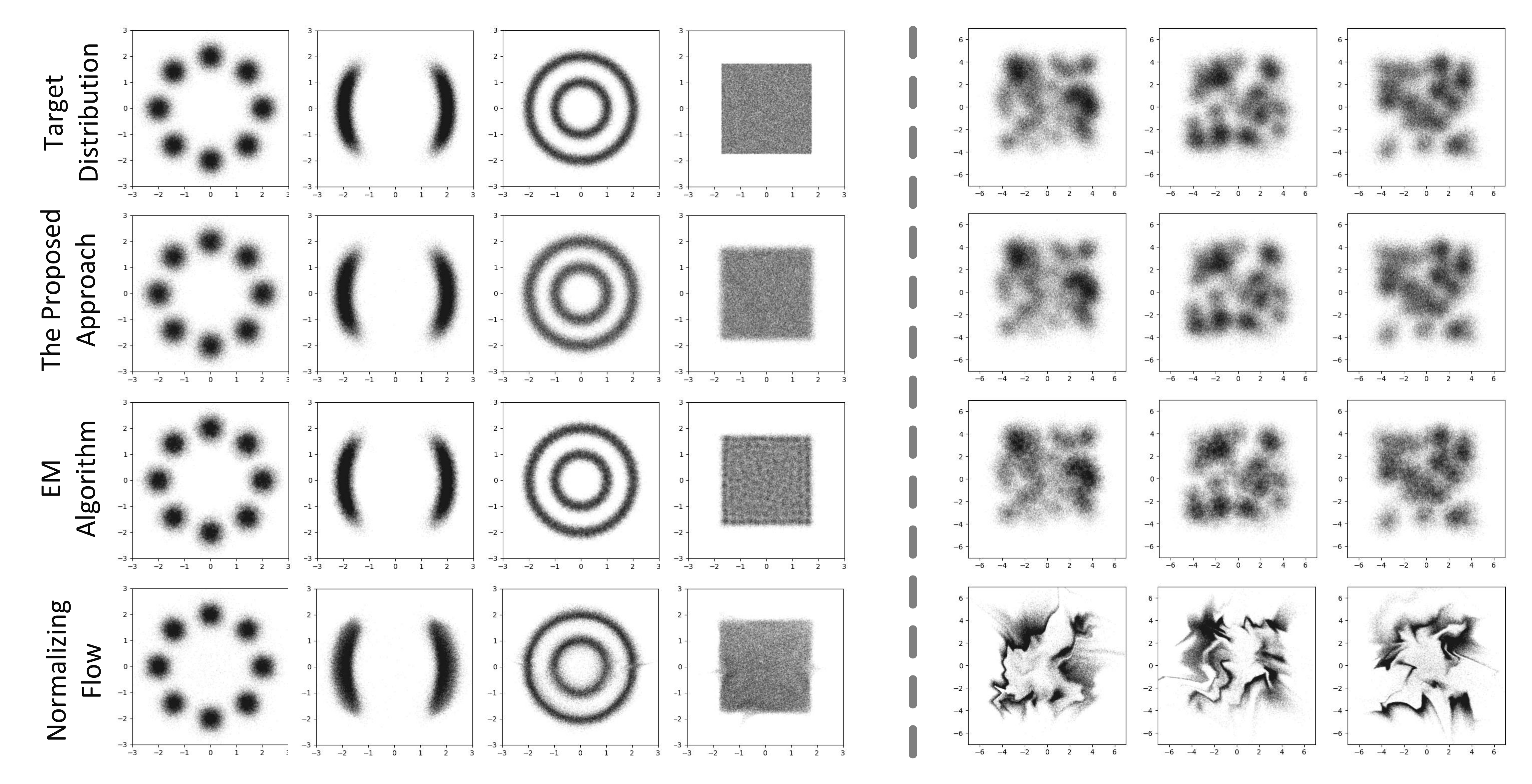}
    \DeclareGraphicsExtensions.
	\caption{Ccomparison of the proposed approach, EM algorithm and Normalizing flow. We first generate a few simple distributions for comparison. Then, complex distribution based on GMM with random parameters are generated for comparison.}
	    \label{fig_discomp}		
\end{figure*}
\subsubsection{Offline Distribution Learning:} Since the optimization of the encoder and the decoder is independent from the distribution learning process, the proposed approach can be devised as an offline learning algorithm, which is useful for saving computation resources.
In the offline learning mode, the training of the proposed disAE can be done in two steps.
In the first step, we can train a regular auto-encoder with mean squared error or cross entropy loss. Once the auto-encoder has been well-trained, we can then learn the distribution based on the proposed Monte-Carlo marginalization.
The process of training an auto-encoder is as follows:
\begin{equation}
	\begin{aligned}
	\alpha, \beta & = \mathop{\text{argmin}}_{\alpha, \beta } \mathbb{E}_{q_{\alpha}(\mathbf{z}|\mathbf{x}) }\text{log}(g_{\beta}(\mathbf{x}|\mathbf{z})) \\
	& =\mathop{\text{argmin}}_{\alpha, \beta } ||\mathbf{x} - g_{\beta}(q_{\alpha}(\mathbf{x}))||_2 
	\end{aligned}
\end{equation}
where, $\alpha$ and $\beta$ denote the parameters of encoder $q()$ and $g()$ respectively.
After parameters of the auto-encoder have been well-updated or updated in one epoch, we can then learn the distribution based on the proposed Monte-Carlo Marginalization, as follows:
\begin{equation}
	\begin{aligned}
		\hat{\boldsymbol{\theta}}  & =  \mathop{\text{argmin}}_{\theta} \mathcal{M}(q(\mathbf{z}), p(\mathbf{z};\boldsymbol{\theta})) \\
& =  \mathop{\text{argmin}}_{\theta} \sum_{\vec{u} \sim \mathbb{U} }  D_{KL}(q_{\vec{u}}(\mathbf{z } \cdot \vec{u} )||p_{\vec{u}}(\mathbf{z} \cdot \vec{u};\theta)).
\end{aligned}
\end{equation}
Note that the distribution learning process can also be applied for a pre-trained VAE. It can produce a distribution generating better image quality compared to the Gaussian distribution which is the fundamental assumption of VAEs.
\section{Experimental results}
Given restrcitions on the length of this paper, we can only briefly discuss the properties of the proposed approach, without introducing details of the experimental setting. Please check the supplementary materials for further details.
All the experiments are evaluated on an RTX A4000 with 16 GB memory. The source code will be available following the acceptance of this paper.
\begin{figure*}[!t]
    \centering
	\includegraphics[width=0.99\linewidth]{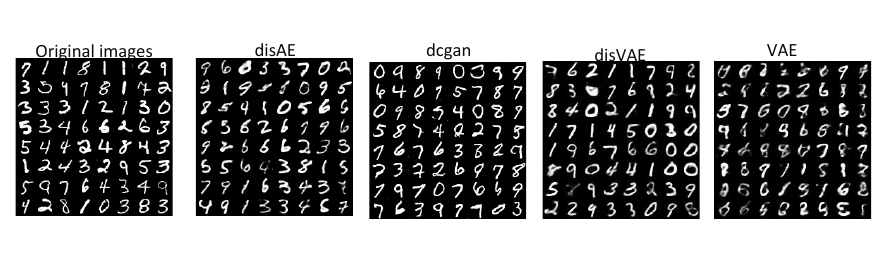}
    \DeclareGraphicsExtensions.
		\caption{Qualitative comparison between disAE, dcGAN, disVAE and VAE. In particular, the results of disVAE is generated based on the VAE's decoder and the distribution learned by the proposed approach.}
	    \label{fig_comp_genimgs}
\end{figure*}
\subsection{Ablation Study}
% With enough Gaussian components, GMM can approximate almost any distribution. 
% Thus, the proposed approach has excellent ability to handle complex distributions.
% In order to demonstrate this, we propose an ablation study on the number of Gaussian components as shown in \reffig{fig_gmm_num}.
% In particular, the learned GMM becomes more and more close to the target distribution when the number of Gaussian components increase.
The proposed approach is differentiable in the entire process. This means that the  parameters of GMM can be updated based on the samples, and the values of the samples can also be updated based on GMM. 
In order to demonsrate this, we first update the GMM based on the samples from the target distribution. 
After the GMM is well-learned, we randomly re-initialize the samples' values and use the learned GMM as target to update the values of the samples.
The processes of updating GMM and samples are shown in \reffig{fig_bidirect}.
Usually, we are more concerned about updating the GMM. 
However, the gradient of samples is also important in the proposed disAE, since the encoder needs gradients from samples with respect to the target distribution to perform backpropagation.
Differentiability is the biggest advantage of the proposed approach compared to the EM algorithm.

\subsection{Learning Intractable Distributions}
The EM algorithm is the most closely related to the proposed approach and normalizing flow is also somewhat related to our work.
Thus, we compare the proposed approach with EM and normalizing flow.
For the EM algorithm, we use the implementation in the Scikit-learn package \cite{pedregosa2011scikit}.
For normalizing flow, we utilize the implmentation from a state-of-the-art work \cite{2022_PMLR_stimper22a}.
The comparison results are shown in \reffig{fig_discomp}.

As shown in \reffig{fig_discomp}, for simple distributions the performance of the proposed approach, EM algorithm and normalizing flow are quite promissing.
However, given the convexity assumption of EM algorithms, the distributions of EM is not smooth. 
This disadvantage becomes more obvious considering a uniform distribution (the distribution in the fourth column).
When the target distributions become complex, the normalizing flow no longer works. 
This may be because the number of iterations for normalizing flow is not enough, or the number of basic distributions such as Gaussian distributions for normalizing flow is not enough.
However, both the EM algorithm and the proposed approach still works pretty well.
As EM algorihtms are one of the best methods devised to find the parameters of GMM, the efficiency of the proposed approach is demonstrated.
In addition, EM algorithms involve iteration steps. Thus, it is hard to integrate them with backpropagation algorithms. 
Thus, the advantage of the proposed approach is demonstrated, given its differentiability.

\subsection{Distribution Learning Auto-Encoder}
In order to show strong evidence that the proposed disAE is better than VAE, we apply the proposed approach into a pre-trained VAE for comparison.
Quantitative and qualitative comparisons are shown in \reftab{tab_fid} and \reffig{fig_comp_genimgs} respectively.
During the comparisons, we download the pre-trained VAE from Github\footnote{https://github.com/csinva/gan-vae-pretrained-pytorch} to generate the images of VAE whose Fid value is 40.1. (Note that we did not contribute to this GitHub link.)
We utilize the proposed Monte-Carlo marginalization method to learn the distribution of the latent variables of the pre-trained VAE. Then, we generate new images based on the distributions learned by the proposed approach to compute the Fid value of disVAE, which is 20.9.
This experiment demonstrates two points: a) There is a gap between true distributions of latent variables and Gaussian distributions. 
b) The proposed approach is able to learn a distribution which is closer to the true distribution compared to Gaussian.
Next, we completely remove the variational inference part of VAE and incorporate the proposed Monte-Carlo marginalization to create a disAE which is re-trained based on the same dataset.
The Fid value of disAE decreases to 10.4 which is even lower than the one of dcGAN, which has 7 times the number of parameters and 3 times the number of dimensions.
The efficiency of the proposed approach is thus demonstrated.
\begin{table}[ht!]
	\caption{Evaluation of the proposed frequency regularization on Generative Adversarial Network (GAN) and Variational Autoencoder (VAE).}
		\setlength{\tabcolsep}{0.8em}
	\label{tab_fid}
		\begin{tabular}{l|rrrr@{}}
			\toprule
			\makecell{} & \makecell{VAE} & \makecell{disVAE} &  \makecell{disAE}	& \makecell{dcGAN}	 	 		\\ 
			\midrule
			\midrule
						Fid Value	& \makecell{40.1}	& \makecell{20.9} 	&	\makecell{10.4}	 &\makecell{10.9} 	 			\\
			\midrule
			NoF	& \makecell{0.6M}	& \makecell{0.6M} 	&	\makecell{0.6M}	 &\makecell{4.2M} 		\\
			\midrule
			NoD & \makecell{20}		& \makecell{20}		&	\makecell{20}		& \makecell{64}		\\
			\bottomrule
		\end{tabular} \\
		\scriptsize NoF: Number of parameters, \ \ NoD: Number of dimensions in latent space.
		\end{table}
\begin{figure}[!t]
    \centering
    \includegraphics[width=0.99\linewidth]{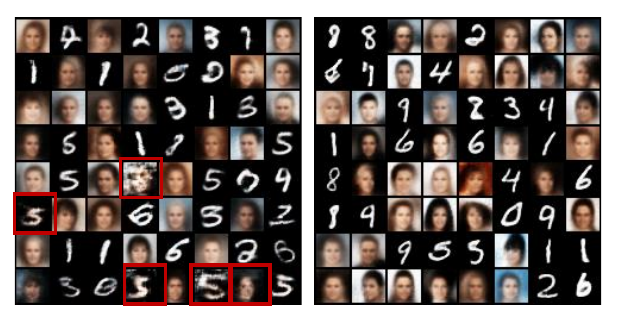}
    \DeclareGraphicsExtensions.
		\caption{Illustration of images generated by VAE (left) and disAE (right). There are a few twisted images in VAE's results, but no twisted images are generated by the proposed disAE.}
    \label{fig_aevae}		
	\end{figure}
\subsection{Discussion and Limitations}
	Compared to VAE, the biggest advantage the proposed disAE is that the gap between a true distribution and the learned distribution is very small.
	Consequently, the proposed disAE rarely generates twisted images. 
	In order to demonstrate this, we trained VAE and disAE on images from MNIST and CelebA.
	A few images generated by VAE and disAE are shown in \reffig{fig_aevae}.
	As shown in \reffig{fig_aevae}, there are a few twisted images (highlighted with red rectangles) in VAE's results.
	By comparison, there are almost no twisted images generated by the proposed disAE.

	The proposed approach also has some disadvatanges. 
	Although the proposed Monte-Carlo marginalization reduces the computational complexity of KL-divergence in high dimension spaces, there is a trade-off between run time and memory requirements, which needs more training epochs.
	However, benefitting from todays techniques on GPU computation, such cost is quite acceptable. 
	A complex 2D distribution can be learned within a few minutes, which is faster than implementations of EM in the Scikit-learn package \cite{pedregosa2011scikit}.
	We demonstrate this through several demo videos in the supplementary material.

\section{Conclusion}
	We proposed the Monte-Carlo Marginalization method to learn the parameters of Gaussian Mixture Models (GMMs) to approximate intractable distributions.
	In the proposed approach, we first marginalize a GMM and the intractable distribution on a random unit vector to compute the their marginal distributions.
	By minimizing the KL-divergence between marginal distributions, GMM graduately approximated the intractable distribution.
	In particular, Kernel density estimation was used to capture the probability density function of intractable distributions, 
	and the marginal distribution of GMM was computed based on its parameters.
	With a sufficient number Gaussian components, the proposed approach can learn almost any intractable distribution. Thus, the distribution learning Auto-Encoder (disAE) was proposed, and its superiority was demonstrated through comparisons with the variational Auto-Encoder (VAE).
\bibliography{aaai23}

\begin{thebibliography}{32}
\providecommand{\natexlab}[1]{#1}

\bibitem[{Bond-Taylor et~al.(2022)Bond-Taylor, Leach, Long, and
  Willcocks}]{2022_TPAMI_9555209}
Bond-Taylor, S.; Leach, A.; Long, Y.; and Willcocks, C.~G. 2022.
\newblock Deep Generative Modelling: A Comparative Review of VAEs, GANs,
  Normalizing Flows, Energy-Based and Autoregressive Models.
\newblock \emph{IEEE Transactions on Pattern Analysis and Machine
  Intelligence}, 44(11): 7327--7347.

\bibitem[{Croitoru et~al.(2023)Croitoru, Hondru, Ionescu, and
  Shah}]{2023_TPAMI_10081412}
Croitoru, F.-A.; Hondru, V.; Ionescu, R.~T.; and Shah, M. 2023.
\newblock Diffusion Models in Vision: A Survey.
\newblock \emph{IEEE Transactions on Pattern Analysis and Machine
  Intelligence}, 1--20.

\bibitem[{Dempster, Laird, and Rubin(1977)}]{EM_477e7e2b}
Dempster, A.~P.; Laird, N.~M.; and Rubin, D.~B. 1977.
\newblock Maximum Likelihood from Incomplete Data via the EM Algorithm.
\newblock \emph{Journal of the Royal Statistical Society. Series B
  (Methodological)}, 39(1): 1--38.

\bibitem[{Dinh, Sohl-Dickstein, and Bengio(2016)}]{dinh2016density}
Dinh, L.; Sohl-Dickstein, J.; and Bengio, S. 2016.
\newblock Density estimation using Real NVP.
\newblock In \emph{International Conference on Learning Representations}.

\bibitem[{et~al.(2019)}]{paszke2019pytorch}
et~al., P. 2019.
\newblock Pytorch: An imperative style, high-performance deep learning library.
\newblock In \emph{Advances in neural information processing systems}.

\bibitem[{Feller(1991)}]{feller1991introduction}
Feller, W. 1991.
\newblock \emph{An introduction to probability theory and its applications,
  Volume 2}, volume~81.
\newblock John Wiley \& Sons.

\bibitem[{Gao et~al.(2017)Gao, Xing, Xie, Wu, and Geng}]{7890384}
Gao, B.-B.; Xing, C.; Xie, C.-W.; Wu, J.; and Geng, X. 2017.
\newblock Deep Label Distribution Learning With Label Ambiguity.
\newblock \emph{IEEE Transactions on Image Processing}, 26(6): 2825--2838.

\bibitem[{Gao et~al.(2021)Gao, Song, Poole, Wu, and Kingma}]{gao2021learning}
Gao, R.; Song, Y.; Poole, B.; Wu, Y.; and Kingma, D. 2021.
\newblock Learning energy-based models by diffusion recovery likelihood.
\newblock In \emph{International Conference on Learning Representations (ICLR
  2021)}.

\bibitem[{Goodfellow, Bengio, and Courville(2016)}]{goodfellow2016deep}
Goodfellow, I.; Bengio, Y.; and Courville, A. 2016.
\newblock \emph{Deep learning}.
\newblock MIT press.

\bibitem[{Hastings(1970)}]{hastings1970monte}
Hastings, W.~K. 1970.
\newblock Monte Carlo sampling methods using Markov chains and their
  applications.

\bibitem[{Hung and Hsieh(2023)}]{hung2023reward}
Hung, Y.-H.; and Hsieh, P.-C. 2023.
\newblock Reward-Biased Maximum Likelihood Estimation for Neural Contextual
  Bandits: A Distributional Learning Perspective.
\newblock In \emph{Proceedings of the AAAI Conference on Artificial
  Intelligence}, volume~37, 7944--7952.

\bibitem[{Kearns et~al.(1994)Kearns, Mansour, Ron, Rubinfeld, Schapire, and
  Sellie}]{kearns1994learnability}
Kearns, M.; Mansour, Y.; Ron, D.; Rubinfeld, R.; Schapire, R.~E.; and Sellie,
  L. 1994.
\newblock On the learnability of discrete distributions.
\newblock In \emph{Proceedings of the twenty-sixth annual ACM symposium on
  Theory of computing}, 273--282.

\bibitem[{Kingma and Ba(2015)}]{kingma2014adam}
Kingma, D.~P.; and Ba, J. 2015.
\newblock Adam: A method for stochastic optimization.
\newblock In \emph{Proc. International Conference on Learning Representations}.

\bibitem[{Kingma and Dhariwal(2018)}]{kingma2018glow}
Kingma, D.~P.; and Dhariwal, P. 2018.
\newblock Glow: Generative flow with invertible 1x1 convolutions.
\newblock \emph{Advances in neural information processing systems}, 31.

\bibitem[{Kingma and Welling(2014)}]{2014_VAE_Kingma2014}
Kingma, D.~P.; and Welling, M. 2014.
\newblock Auto-Encoding Variational Bayes.
\newblock In \emph{International Conference on Learning Representations, 2014}.

\bibitem[{LeCun(1998)}]{1998_lecun1998mnist}
LeCun, Y. 1998.
\newblock The MNIST database of handwritten digits.
\newblock \emph{http://yann. lecun. com/exdb/mnist/}.

\bibitem[{LeCun et~al.(2006)LeCun, Chopra, Hadsell, Ranzato, and
  Huang}]{lecun2006tutorial}
LeCun, Y.; Chopra, S.; Hadsell, R.; Ranzato, M.; and Huang, F. 2006.
\newblock A tutorial on energy-based learning.
\newblock \emph{Predicting structured data}, 1(0).

\bibitem[{Liang, An, and Ma(2022)}]{2022_aaai_Liang}
Liang, J.; An, P.; and Ma, J. 2022.
\newblock Distribution Aware VoteNet for 3D Object Detection.
\newblock \emph{Proceedings of the AAAI Conference on Artificial Intelligence},
  36(2): 1583--1591.

\bibitem[{Liu et~al.(2023)Liu, Zhang, Zhang, Zhao, Ma, Wu, Chen, Yu, and
  Zhang}]{liu2023boosting}
Liu, H.; Zhang, F.; Zhang, X.; Zhao, S.; Ma, F.; Wu, X.-M.; Chen, H.; Yu, H.;
  and Zhang, X. 2023.
\newblock Boosting Few-Shot Text Classification via Distribution Estimation.
\newblock \emph{arXiv preprint arXiv:2303.16764}.

\bibitem[{Liu et~al.(2015)Liu, Luo, Wang, and
  Tang}]{2015_celeba_liu2015faceattributes}
Liu, Z.; Luo, P.; Wang, X.; and Tang, X. 2015.
\newblock Deep Learning Face Attributes in the Wild.
\newblock In \emph{Proceedings of International Conference on Computer Vision
  (ICCV)}.

\bibitem[{Lu et~al.(2023)Lu, Ning, Liu, Sun, Zhang, Yang, and Wang}]{lu2023opt}
Lu, M.; Ning, S.; Liu, S.; Sun, F.; Zhang, B.; Yang, B.; and Wang, L. 2023.
\newblock OPT-GAN: A Broad-Spectrum Global Optimizer for Black-Box Problems by
  Learning Distribution.
\newblock In \emph{Proceedings of the AAAI Conference on Artificial
  Intelligence}, volume~37, 12462--12472.

\bibitem[{Papamakarios, Pavlakou, and Murray(2017)}]{papamakarios2017masked}
Papamakarios, G.; Pavlakou, T.; and Murray, I. 2017.
\newblock Masked autoregressive flow for density estimation.
\newblock \emph{Advances in neural information processing systems}, 30.

\bibitem[{Pedregosa et~al.(2011)Pedregosa, Varoquaux, Gramfort, Michel,
  Thirion, Grisel, Blondel, Prettenhofer, Weiss, Dubourg
  et~al.}]{pedregosa2011scikit}
Pedregosa, F.; Varoquaux, G.; Gramfort, A.; Michel, V.; Thirion, B.; Grisel,
  O.; Blondel, M.; Prettenhofer, P.; Weiss, R.; Dubourg, V.; et~al. 2011.
\newblock Scikit-learn: Machine learning in Python.
\newblock \emph{the Journal of machine Learning research}, 12: 2825--2830.

\bibitem[{Qiang et~al.(2023)Qiang, Hou, Wan, Liang, Lei, and
  Zhang}]{qiang2023mixture}
Qiang, S.; Hou, J.; Wan, J.; Liang, Y.; Lei, Z.; and Zhang, D. 2023.
\newblock Mixture Uniform Distribution Modeling and Asymmetric Mix Distillation
  for Class Incremental Learning.

\bibitem[{Stimper, Sch\"olkopf, and Miguel
  Hernandez-Lobato(2022)}]{2022_PMLR_stimper22a}
Stimper, V.; Sch\"olkopf, B.; and Miguel Hernandez-Lobato, J. 2022.
\newblock Resampling Base Distributions of Normalizing Flows.
\newblock In Camps-Valls, G.; Ruiz, F. J.~R.; and Valera, I., eds.,
  \emph{Proceedings of The 25th International Conference on Artificial
  Intelligence and Statistics}, volume 151 of \emph{Proceedings of Machine
  Learning Research}, 4915--4936. PMLR.

\bibitem[{Tabak and Turner(2013)}]{tabak2013family}
Tabak, E.~G.; and Turner, C.~V. 2013.
\newblock A family of nonparametric density estimation algorithms.
\newblock \emph{Communications on Pure and Applied Mathematics}, 66(2):
  145--164.

\bibitem[{Vuckovic(2022)}]{NEURIPS2022_b6341525}
Vuckovic, J. 2022.
\newblock Nonlinear MCMC for Bayesian Machine Learning.
\newblock In Koyejo, S.; Mohamed, S.; Agarwal, A.; Belgrave, D.; Cho, K.; and
  Oh, A., eds., \emph{Advances in Neural Information Processing Systems},
  volume~35, 28400--28413. Curran Associates, Inc.

\bibitem[{Yu et~al.(2023)Yu, Yu, Song, Neiswanger, and Ermon}]{yu2023offline}
Yu, L.; Yu, T.; Song, J.; Neiswanger, W.; and Ermon, S. 2023.
\newblock Offline imitation learning with suboptimal demonstrations via relaxed
  distribution matching.
\newblock In \emph{Proceedings of the AAAI conference on artificial
  intelligence}, volume~37, 11016--11024.

\bibitem[{Zhang, Li, and Li(2020)}]{2020_aaai_policy}
Zhang, C.; Li, Y.; and Li, J. 2020.
\newblock Policy Search by Target Distribution Learning for Continuous Control.
\newblock \emph{Proceedings of the AAAI Conference on Artificial Intelligence},
  34(04): 6770--6777.

\bibitem[{Zhang, Zhao, and Jin(2020)}]{2020_aaai_optimal}
Zhang, T.; Zhao, P.; and Jin, H. 2020.
\newblock Optimal Margin Distribution Learning in Dynamic Environments.
\newblock \emph{Proceedings of the AAAI Conference on Artificial Intelligence},
  34(04): 6821--6828.

\bibitem[{Zhao et~al.(2023)Zhao, An, Xu, Wang, and Geng}]{zhao2023imbalanced}
Zhao, X.; An, Y.; Xu, N.; Wang, J.; and Geng, X. 2023.
\newblock Imbalanced Label Distribution Learning.
\newblock In \emph{Proceedings of the AAAI Conference on Artificial
  Intelligence}, volume~37, 11336--11344.

\bibitem[{Zhou et~al.(2021)Zhou, Huang, Zhou, and Shao}]{2021_aaai_Zhou}
Zhou, Y.; Huang, L.; Zhou, T.; and Shao, L. 2021.
\newblock Many-to-One Distribution Learning and K-Nearest Neighbor Smoothing
  for Thoracic Disease Identification.
\newblock \emph{Proceedings of the AAAI Conference on Artificial Intelligence},
  35(1): 768--776.

\end{thebibliography}

\end{document}